\documentclass[conference]{IEEEtran}
\IEEEoverridecommandlockouts
\usepackage[style=ieee]{biblatex} 
\bibliography{example_bib.bib}    
\usepackage{amssymb}
\usepackage{amsmath}
\usepackage{algorithmicx,algorithm, algpseudocode}
\usepackage{amsmath}
\usepackage{multirow}
\usepackage{textcomp}
\usepackage{xcolor}
\usepackage{biblatex} 
\usepackage{tabularx}
\usepackage{amsmath}

\usepackage{url}
\usepackage{graphicx}  
\usepackage{float}  

\usepackage[colorlinks]{hyperref}

\newcommand{\linebreakand}{%
    \end{@IEEEauthorhalign}
    \hfill\mbox{}\par
    \mbox{}\hfill\begin{@IEEEauthorhalign}
}

\makeatother
\graphicspath{ {./images/} }

\begin{document}

\title{Online pseudo labeling for polyp segmentation with momentum networks \\}

\author{\IEEEauthorblockN{Toan Pham Van}
    \IEEEauthorblockA{\textit{Sun Asterisk R\&D Unit} \\
    pham.van.toan@sun-asterisk.com}
    \and
    \IEEEauthorblockN{Linh Bao Doan}
    \IEEEauthorblockA{\textit{Sun Asterisk R\&D Unit} \\
    doan.bao.linh@sun-asterisk.com}
    \and
    \IEEEauthorblockN{Thanh Tung Nguyen}
    \IEEEauthorblockA{\textit{Sun Asterisk R\&D Unit} \\
    nguyen.tung.thanh@sun-asterisk.com}
    \and
    \IEEEauthorblockN{Duc Trung Tran}
    \IEEEauthorblockA{\textit{Sun Asterisk R\&D Unit} \\
    tran.trung.duc@sun-asterisk.com}
    \and
    \IEEEauthorblockN{Quan Van Nguyen}
    \IEEEauthorblockA{\textit{Sun Asterisk R\&D Unit} \\
    nguyen.van.quan@sun-asterisk.com}
    \and 
    \IEEEauthorblockN{Dinh Viet Sang}
    \IEEEauthorblockA{\textit{Hanoi University of Science and Technology} \\
    sangdv@soict.hust.edu.vn}

}


\maketitle

\begin{abstract} 
Semantic segmentation is an essential task in developing medical image diagnosis systems. However, building an annotated medical dataset is expensive. Thus, semi-supervised methods are significant in this circumstance. In semi-supervised learning, the quality of labels plays a crucial role in model performance. In this work, we present a new pseudo labeling strategy that enhances the quality of pseudo labels used for training student networks. We follow the multi-stage semi-supervised training approach, which trains a teacher model on a labeled dataset and then uses the trained teacher to render pseudo labels for student training. By doing so, the pseudo labels will be updated and more precise as training progress. The key difference between previous and our methods is that we update the teacher model during the student training process. So the quality of pseudo labels is improved during the student training process. We also propose a simple but effective strategy to enhance the quality of pseudo labels using a momentum model - a slow copy version of the original model during training. By applying the momentum model combined with re-rendering pseudo labels during student training, we achieved an average of 84.1\% Dice Score on five datasets (i.e., Kvarsir, CVC-ClinicDB, ETIS-LaribPolypDB, CVC-ColonDB, and CVC-300) with only 20\% of the dataset used as labeled data. Our results surpass common practice by 3\% and even approach fully-supervised results on some datasets. Our source code and pre-trained models are available at \textit{\url{https://github.com/sun-asterisk-research/online_learning_ssl}}
\end{abstract}


\section{Introduction}


A polyp is a grossly visible mass of epithelial cells that protrudes from the mucosal surface into the lumen of the intestine. Most colon polyps are harmless. However, some colon polyps can develop into colon cancer over time, which is fatal when found in its late stages. Hence, it is essential to have regular screenings. Colonoscopy is one of the most common techniques used to detect colon polyps. With the recent success of deep learning, various image segmentation methods have been employed to help oncologists save significant time. However, deep learning models often require a large amount of labeled data. In real-world scenarios, this can be difficult, especially for medical data. Producing a labeled medical dataset for training semantic segmentation models requires the laborious creation of pixel-level labels and annotators' expertise and medical knowledge in medical image diagnosis.

Semi-supervised learning is a method that utilizes a massive amount of unlabeled data for training deep neural networks on actual tasks. This method is significant with medical image data since the cost of labeling the data is often high and requires much effort from many experts. Many studies applied semi-supervised learning on medical image data, such as pseudo labeling \cite{wang2022semi}, cross pseudo supervision \cite{zhang2022semi}, few-shot learning \cite{li2021few}, deep adversarial learning, and so on. These methods use both labeled and unlabeled data to train deep learning models for the semantic segmentation task. Most of them aim to generate high-quality pseudo labels and have more robust representations of the domain distribution of the data. Previous works \cite{araslanov2021self} have shown the effectiveness of the momentum network - a slow copy version of the training model by taking an exponential moving average. Empirical results show that the momentum network often produces more stable results than the original model trained directly through the back-propagation process.

In current semi-supervised learning methods, pseudo label generation can be done online as in \cite{araslanov2021self} or offline as in \cite{yang2020fda, zou2019confidence}. The advantage of online pseudo labeling is that it is simple and easy to implement with only one training stage. In addition, this method will be effective when the pseudo label generation model is updated online to help enhance the quality of pseudo labels after each iteration. This paper combines online pseudo labeling and momentum networks to improve the quality of pseudo labels generated after each iteration. This method improves Dice Score by 3\% compared with offline pseudo labeling and outperforms supervised models in some out-of-domain datasets.

Our main contributions in this paper are:
\begin{itemize}
\item We propose a novel training strategy for semi-supervised learning which combines online pseudo labeling and momentum networks for the polyp segmentation task. Our results on Dice Score are about 3\% better than the offline pseudo labeling method and outperform supervised models in some out-of-domain datasets.
\item Comprehensive analysis of the effects of online pseudo labeling and momentum network on our results. 
\item To the best of our knowledge, this is the first study to combine online pseudo labeling and momentum network for polyp segmentation task.
\end{itemize}

The rest of this paper is organized as follows. Related work is presented in Section 2. The proposed method is presented in Section 3. Section 4 describes the datasets and details implementation methods. The experimental results are presented in Section 5. Finally, discussions and future research directions are presented in Section 6.

\section{Related Work}
\subsection{Semantic Segmentation}

Semantic segmentation aims to classify images at the pixel level. Since the Fully Convolutional Neural Network was applied to semantic segmentation in the study of Long et al. \cite{long2015fully}, deep learning models have achieved remarkable results in this field. Currently, most segmentation models are designed based on the encoder-decoder architecture \cite{long2015fully}. A few studies also aim to build a specific model for polyp segmentation. HarDNet-MSEG \cite{huang2021hardnet} achieved high performance on the Kvasir-SEG dataset with processing speed up to 86 FPS. AG-ResUNet++ improved UNet++ with attention gates and ResNet backbone. Another study called Transfuse combined Transformer and CNN using BiFusion module \cite{zhang2021transfuse}. ColonFormer \cite{duc2022colonformer} used MiT backbone, Upper decoder, and residual axial reverse attention to further boost the polyp segmentation accuracy. NeoUNet \cite{ngoc2021neounet} and BlazeNeo \cite{an2022blazeneo} proposed effective encoder-decoder networks for polyp segmentation and neoplasm detection. Generally, research in changing model architecture is still a potential approach. However, this approach often requires a large amount of labeled data. Collecting dense pixel-level annotations for semantic segmentation is costly and difficult, especially for medical segmentation data. Therefore, applying an efficient learning method to help deal with the lack of data is critical for practical.

\subsection{Semi supervised learning for semantic segmentation}

One of the disadvantages of supervised learning is the requirement of massive amounts of labeled data. This problem becomes even more difficult with medical image data such as polyp segmentation because it requires effort from people with experience and expertise in imaging diagnostic. Semi-supervised learning solves this problem by combining labeled and unlabelled data during the training model. Several studies have applied semi-supervised learning to semantic segmentation problems, such as adversarial learning, consistency regularization, or pseudo-labeling \cite{feng2020semi}. General objective of semi-supervised learning is improving base model trained on dataset of $N$ labelled samples $D_{sup}=\left\{ (x_n, y_n) | n=1...N \right\}$ by utilizing a dataset of M unlabeled samples $D_{unsup}=\left\{ x_m | m=N+1...N+M \right\}$. 

In this paper, we focus on improving the quality of pseudo labels by combining the online pseudo labeling strategy with momentum networks. 

\subsection{Momentum Network}

A momentum network is a slow copy version of weights of the original model during the training process through Exponential Moving Average (EMA). The update process can be done after each epoch or after a certain number of steps in the training process. 

$$\theta'_t=\alpha\theta'_{t-1} + (1 - \alpha)\theta_t$$
where $\theta'_t$ and $\theta_t$ are the weights of the momentum and original models, respectively, at the $t$-th step during the training process. The studies applying momentum network have shown experimentally that momentum network gives more stable accuracy than the original model \cite{araslanov2021self}. Accordingly, the momentum model can be considered an ensemble of several versions of the original model at different time steps during training.

\subsection{Online vs. offline pseudo label generation}

Pseudo-label generation methods can be divided into two types: online generation and offline generation. Online generation generates pseudo labels directly during the forwarding process \cite{sohn2020fixmatch}. The offline generation will generate pseudo labels once in each semi-supervised training, and pseudo labels will only change in the subsequent iterations \cite{iscen2019label, chen2020big}. The advantage of offline learning is that the quality of the pseudo label will not change in each iteration, and the more iterations we train, the better the quality of the pseudo label will be. However, offline pseudo label generation requires more complicated installation and computational costs than online pseudo label generation. The installation is straightforward with the online pseudo label generation method and only requires one-stage training with the student model. However, the online generation method demands a stable quality of the pseudo label during the label generation process. That needs the pseudo-label generation model to be smooth. We solve this problem by using a momentum network during training semi-supervised learning.  

\section{Proposed method}

In this section, we present an online prototyping strategy that ensures the stability of fake label quality during prototyping with a combination of online pseudo labeling and momentum networks. Details of the available pipeline, loss function configurations as well as augmentation methods for each data type are also detailed in this section.

\subsection{Overall pipeline}

\begin{figure*}[ht]
\centering
\includegraphics[width=0.8\linewidth]{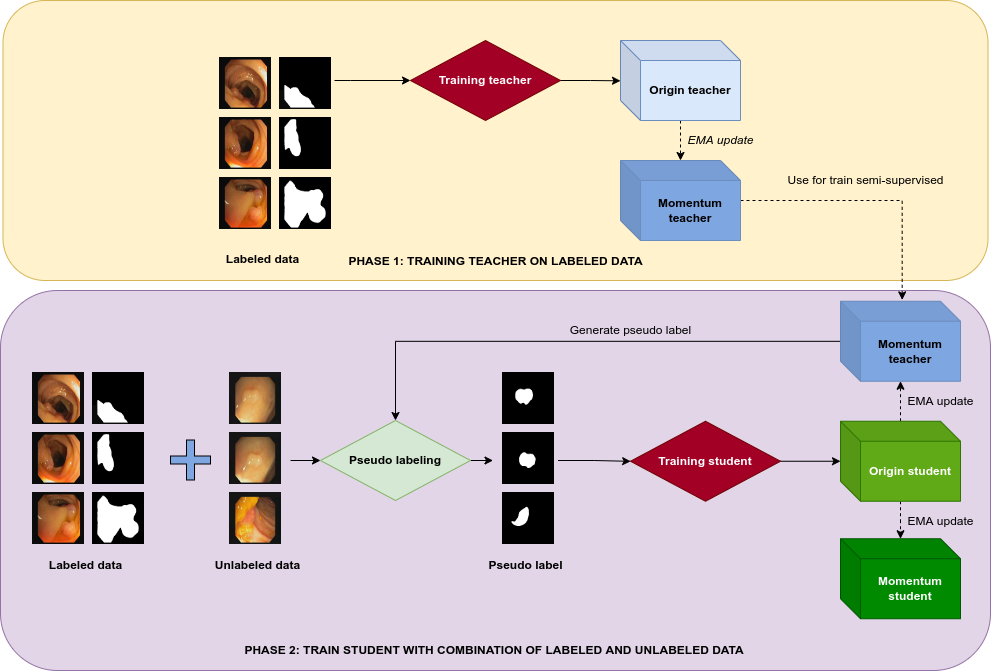}
\caption{Overview of online pseudo labeling with momentum network pipeline}
\label{fig:pipeline}
\end{figure*}

The overall architecture of the system is described in Fig.~\ref{fig:pipeline}. We still apply a two-stage strategy for semi-supervised training, where the teacher model is first trained on the labeled dataset. During the training process, both the original model and the slow copy version of the model updated by EMA (Momentum teacher network) are saved and served for student training. We perform the online pseudo label generation during the student training process using the momentum teacher model. At the same time, the weights of the original student model are updated for the momentum teacher and its student momentum version. Finally, the momentum student version will be used to predict the outcome after the training finishes. Details of each training step will be described in the sections below.

\subsubsection{Teacher training}
The training teacher model is performed on the entire labeled dataset. We use the Tversky loss function during training. Besides, we use the FPN architecture with the DenseNet169 backbone as the primary model for our experiment. The best momentum teacher network on the validation set will be used to generate pseudo labels during student training.
\subsubsection{Student training}

After having the momentum teacher model trained on the dataset labeled $D_{sup}$, we proceed to train the student model with the pseudo label generated from the momentum teacher model. The difference is that the weights of the momentum teacher model will also be updated during training after a certain number of training steps. This strategy helps the fake labels to be constantly updated and stable due to the nature of the momentum network. The detailed algorithm is presented in \textbf{Algorithm} \ref{algo:main_stategy}.

\begin{algorithm}
\caption{Online Pseudo Labeling with Momentum Teacher Algorithm}
\label{algo:main_stategy}
\textbf{Input}: Labeled images $D_{sup}$, Unlabeled images $D_{unsup}$, Trained momentum teacher $MT_0$ \\
\textbf{Output}: Best momentum student $MS_{best}$ trained with combination of supervised and unsupervised data 
\begin{algorithmic}
\Procedure{OnlineLabelingMomentum}{}

\For {$t=0$ to $n\_epochs$}
\State \textbf{Step 1:} Generate pseudo labels of $D_{unsup}$ with $MT_t$ model.
\State \textbf{Step 2:} Train student $S_t$ with combination of $D_{sup}$ and $D_{unsup}$ with generated pseudo labels.
\State \textbf{Step 3:} Update momentum teacher $MT_{t}$ with $S_t$ weights via EMA.
\State \textbf{Step 3:} Update momentum student $MS_{t}$ with $S_t$ weights via EMA.
\State \textbf{Step 4:} Save the best models in validation set.
\EndFor
\EndProcedure
\end{algorithmic}
\end{algorithm}

\subsection{Data Augmentation}

\subsubsection{Weak Augmentation on labeled dataset}
We choose two different data augmentation strategies for labeled and unlabelled data. We apply the weak or basic augmentation methods to the labeled dataset. For the weak augmentation, the training images are random flips with a probability of 0.5. 
\subsubsection{Strong Augmentation on unlabeled dataset}

When applied to unlabeled data, noise has the essential
benefit of enforcing invariance in the decision function on
both labeled and unlabeled data. When the student is deliberately noised, it is trained to be consistent with the teacher that is not noised when it generates pseudo labels. We only use the input noise in our experiments via a strong augmentation method. We use the combination of ShiftScaleRotate, RGBShift, RandomBrightnessContrast, and RandomFlip with the probability of 0.5.

\section{Dataset and Experiments Setup}

\subsection{Dataset}
To compare the method results, we use the same Polyp dataset as in the study of HardDet-MSEG \cite{huang2021hardnet}. This dataset includes 1450 endoscopic images of size $384 \times 288 \times 3$ corresponding to 1450 masks of size $384 \times 288 \times 1$. The mask is a binary image, the pixels have a value of 255 corresponding to the area containing the polyp, and the pixels in background areas have a value of 0. The training dataset includes 900 images in Kvasir-SEG
and 550 in the CVC-ClinicDB. The test dataset includes 798 images synthesized from different data sets, which are the CVC-300 dataset, CVC-ClinicDB, CVC-ColonDB, ETIS- LaribPolypDB and
Kvasir-SEG. Our experiment uses a 10\% dataset equivalent to 145 images as the validation dataset. The remaining total of 1305 images was used to scale the labeled and unlabeled datasets in each experiment.

\subsection{Evaluation Metrics}
\textbf{Mean IoU} and \textbf{Mean Dice} are used to evaluate the model and compare our experiments. There are defined as follows:

\begin{raggedleft}
\begin{tabular}{p{4cm}p{3.5cm}}
  $$mIoU = \frac{TP}{TP + FP + FN}$$
  &
  $$mDice = \frac{2*TP}{2*TP + FP + FN}$$
\end{tabular}
\end{raggedleft}

\subsection{System configuration}
Our experiments are conducted on a computer with Intel Core i5-7500 CPU @3.4GHz, 32GB of RAM, GPU GeForce GTX 1080 Ti, and 1TB SSD hard disk. The models are implemented with the PyTorch Lightning framework.

\section{Result of Experiments}
This section presents teacher training results on the labeled dataset and semi-supervised learning with offline and online pseudo-labeling. We use the base model and its momentum network for the above experiments to compare the results. We also divide the dataset with different ratios of labeled data to measure the effect of the amount of labeled data on our method.

\subsection{Training teacher model in a supervised manner in labeled data}

First, we train the model on the labeled set. Original model weights are optimized by Adam optimizer with back-propagation algorithms. The slow copy of the original model is updated simultaneously with the momentum ratio is 0.95. We save the best checkpoint of both models in the validation set during training as the final teacher model. Table \ref{tab:compare_teacher} shows the performance comparison between the original teacher model and its momentum network. We found that the momentum model gives better results than the original model. Therefore, we choose the momentum teacher model as the basis for the semi-supervised experiments presented in the following sections.

\begin{table*}[h]
\centering
\caption{A comparison of the original teacher with the momentum model teacher in different ratios of labeled data}
\def\arraystretch{1.1}

\begin{tabular}{|c|c|cc|cc|cc|cc|cc|cc|}
\hline
\textbf{Ratio of} & \textbf{Momentum} & \multicolumn{2}{c|}{\textbf{CVC-ClinicDB}} & \multicolumn{2}{c|}{\textbf{ETIS}} & \multicolumn{2}{c|}{\textbf{CVC-ColonDB}} & \multicolumn{2}{c|}{\textbf{CVC-300}} & \multicolumn{2}{c|}{\textbf{Kvarsir}} & \multicolumn{2}{c|}{\textbf{Average}} \\ \cline{3-14} 
\textbf{labeled data}     & \textbf{teacher}          & \textbf{mIoU}                 & \textbf{mDice}               & \textbf{mIoU}             & \textbf{mDice}           & \textbf{mIoU}                & \textbf{mDice}               & \textbf{mIoU}              & \textbf{mDice}             & \textbf{mIoU}              & \textbf{mDice}             & \textbf{mIoU}              & \textbf{mDice}             \\ \hline
\multirow{2}{*}{20\%}                       & -                                           & 0.794                & 0.856               & 0.618            & 0.698           & 0.625               & 0.702               & 0.803             & 0.875             & 0.819             & 0.883             & 0.732             & 0.803             \\
                                            & \checkmark                   & 0.792                & 0.854               & 0.616            & 0.699           & 0.616               & 0.693               & 0.800             & 0.875             & 0.820             & 0.884             & 0.730             & 0.801             \\ \hline
\multirow{2}{*}{40\%}                       & -                                           & 0.787                & 0.857               & 0.587            & 0.674           & 0.624               & 0.700               & 0.824             & 0.892             & 0.817             & 0.879             & 0.728             & 0.808             \\
                                            & \checkmark                   & 0.803                & 0.868               & 0.616            & 0.694           & 0.642               & 0.716               & 0.831             & 0.898             & 0.819             & 0.876             & 0.743             & 0.811             \\ \hline
\multirow{2}{*}{60\%}                       & -                                           & 0.835                & 0.889               & 0.610            & 0.677           & 0.645               & 0.721               & 0.820             & 0.885             & 0.842             & 0.894             & 0.751             & 0.814             \\
                                            & \checkmark                   & 0.850                & 0.902               & 0.620            & 0.685           & 0.663               & 0.741               & 0.839             & 0.904             & 0.847             & 0.898             & 0.764             & 0.826             \\ \hline
\end{tabular}
\label{tab:compare_teacher}
\end{table*}

\subsection{Training student with offline pseudo labeling with semi-supervised manner}

We use the momentum teacher model for our semi-supervised learning experiments. The student model has the same architecture as the teacher model. In the offline pseudo labeling strategy, the pseudo labels will be generated directly from the momentum teacher model. The momentum teacher model did not change during training. That means the pseudo-labels for the same image will be the same in all training iterations. The student model kept the original and momentum versions as the teacher training on the labeled dataset.
\subsection{Training student with online pseudo labeling with semi-supervised manner}

\begin{figure*}[ht]
\centering
\includegraphics[width=0.6\linewidth]{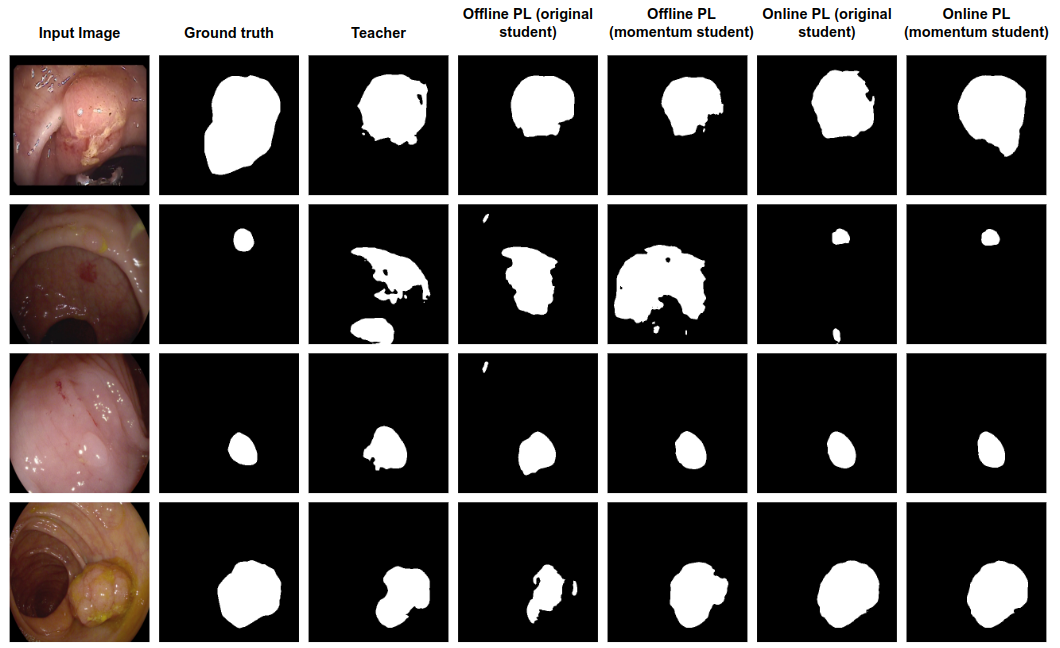}
\caption{Qualitative result comparison between offline pseudo labeling and online pseudo labeling, with/without using momentum network. From left to right: input image in RGB format, ground truth, the output of teacher (momentum network) trained on the labeled dataset, the output of original student trained with offline pseudo labeling, the output of momentum network of original student trained with offline pseudo labeling, the output of original student trained with online pseudo labeling and the output of momentum network of original student trained with online pseudo labeling.}
\label{fig:visualize_result}
\end{figure*}

Online pseudo labeling follows the same training strategy as offline labeling. The only difference is that the Teacher model will also be updated via EMA with the weights of the student model after each epoch. Since then, the pseudo label for an image is also continuously updated during model training. The online learning method combined with the momentum student model gives better results than offline learning. The results are shown in Table \ref{tab:compare_online_offline} and illustrated in Fig. \ref{fig:visualize_result}.

\begin{table*}[h]
\centering
\caption{A comparison of the online pseudo labeling and offline pseudo labeling strategy for semi-supervised training}
\def\arraystretch{1.1}
\resizebox{\textwidth}{!}{
\begin{tabular}{|c|c|c|cc|cc|cc|cc|cc|cc|}
\hline
\textbf{Ratio of}     & \textbf{Online}           & \textbf{Momentum}         & \multicolumn{2}{c|}{\textbf{CVC-ClinicDB}} & \multicolumn{2}{c|}{\textbf{ETIS}} & \multicolumn{2}{c|}{\textbf{CVC-ColonDB}} & \multicolumn{2}{c|}{\textbf{CVC-300}} & \multicolumn{2}{c|}{\textbf{Kvarsir}} & \multicolumn{2}{c|}{\textbf{Average}} \\ \cline{4-15} 
\textbf{labeled data}          & \textbf{pseudo-labels}             & \textbf{student}                   & \textbf{mIoU}                 & \textbf{mDice}               & \textbf{mIoU}             & \textbf{mDice}           & \textbf{mIoU}                & \textbf{mDice}               & \textbf{mIoU}              & \textbf{mDice}             & \textbf{mIoU}              & \textbf{mDice}             & \textbf{mIoU}          & \textbf{mDice}        \\ \hline
\multirow{4}{*}{20\%} & -                         & -                         & 0.789                & 0.847               & 0.590            & 0.670           & 0.647               & 0.727               & 0.821             & 0.891             & 0.832             & 0.891             & 0.736         & 0.805        \\
                      & -                         & \checkmark & 0.830                & 0.887               & 0.676            & 0.754           & 0.676               & 0.755               & 0.828             & 0.897             & 0.843             & 0.897             & 0.770         & 0.838        \\
                      & \checkmark & -                         & 0.801                & 0.857               & 0.673            & 0.748           & 0.641               & 0.717               & 0.830             & 0.897             & 0.835             & 0.895             & 0.756         & 0.823        \\
                      & \checkmark & \checkmark & 0.816                & 0.870               & 0.701            & 0.772           & 0.669               & 0.742               & 0.836             & 0.903             & 0.856             & 0.911             & 0.778         & 0.841        \\ \hline
\multirow{4}{*}{40\%} & -                         & -                         & 0.792                & 0.850               & 0.628            & 0.704           & 0.650               & 0.733               & 0.833             & 0.902             & 0.813             & 0.875             & 0.743         & 0.813        \\
                      & -                         & \checkmark & 0.825                & 0.882               & 0.673            & 0.749           & 0.671               & 0.745               & 0.824             & 0.894             & 0.837             & 0.893             & 0.766         & 0.833        \\
                      & \checkmark & -                         & 0.824                & 0.883               & 0.601            & 0.671           & 0.668               & 0.750               & 0.829             & 0.896             & 0.838             & 0.899             & 0.752         & 0.820        \\
                      & \checkmark & \checkmark & 0.825                & 0.882               & 0.702            & 0.777           & 0.689               & 0.768               & 0.825             & 0.895             & 0.850             & 0.908             & 0.778         & 0.846        \\ \hline
\multirow{4}{*}{60\%} & -                         & -                         & 0.833                & 0.888               & 0.617            & 0.686           & 0.651               & 0.725               & 0.802             & 0.871             & 0.842             & 0.899             & 0.749         & 0.814        \\
                      & -                         & \checkmark & 0.856                & 0.905               & 0.700            & 0.773           & 0.685               & 0.763               & 0.839             & 0.904             & 0.865             & 0.915             & 0.789         & 0.852        \\
                      & \checkmark & -                         & 0.832                & 0.881               & 0.652            & 0.724           & 0.674               & 0.752               & 0.819             & 0.880             & 0.856             & 0.910             & 0.767         & 0.830        \\
                      & \checkmark & \checkmark & 0.855                & 0.901               & 0.694            & 0.772           & 0.701               & 0.777               & 0.833             & 0.898             & 0.865             & 0.916             & 0.790         & 0.853        \\ \hline
\end{tabular}
}
\label{tab:compare_online_offline}
\end{table*}

\subsection{Comparison with some difference supervised methods}

We evaluated our model’s performance with a benchmark consisting of the state-of-the-art models, namely UNet \cite{ronneberger2015u}, UNet++ \cite{zhou2019unet++}, SFA \cite{fang2019selective}, PraNet \cite{pranet}, MSNet \cite{zhao2021automatic} and Shallow Attention \cite{wei2021shallow}. Table \ref{tab:compare_with_supervised} shows our model’s performance results for mIoU and mDice metrics compared to the results of the benchmark studies. Our model outperformed UNet, UNet++, PraNet, SFA, and Shallow Attention on all datasets for Dice and IoU metrics with only using a maximum of 60\% of labeled data. We are only about 2\% mDice worse than MSNET on the CVC-ClinicDB set but much better on the rest of the datasets. In particular, our method gives better generalization. Our method outperforms all above-supervised methods when tested in out-of-distribution datasets such as ETIS-LabribPolypDB, CVC-300, and CVC-ColonDB.

\begin{table*}[h]
\centering
\caption{A comparison of our method with  state-of-the-art supervised models}
\def\arraystretch{1.1}
\begin{tabular}{|c|c|cc|cc|cc|cc|cc|}
\hline
\multirow{2}{*}{\textbf{Methods}}                        & \textbf{Ratio of}     & \multicolumn{2}{c|}{\textbf{CVC-ClinicDB}}       & \multicolumn{2}{c|}{\textbf{ETIS}}               & \multicolumn{2}{c|}{\textbf{CVC-ColonDB}}        & \multicolumn{2}{c|}{\textbf{CVC-300}}            & \multicolumn{2}{c|}{\textbf{Kvarsir}}            \\ \cline{3-12} 
                                                         & \textbf{labeled data} & \textit{\textbf{mIOU}} & \textit{\textbf{mDice}} & \textit{\textbf{mIOU}} & \textit{\textbf{mDice}} & \textit{\textbf{mIOU}} & \textit{\textbf{mDice}} & \textit{\textbf{mIOU}} & \textit{\textbf{mDice}} & \textit{\textbf{mIOU}} & \textit{\textbf{mDice}} \\ \hline
Unet  \cite{ronneberger2015u}           & 100\%                 & 0.755                  & 0.823                   & 0.335                  & 0.398                   & 0.444                  & 0.512                   & 0.627                  & 0.710                   & 0.746                  & 0.818                   \\
Unet++ \cite{zhou2019unet++}            & 100\%                 & 0.729                  & 0.794                   & 0.344                  & 0.401                   & 0.410                  & 0.483                   & 0.624                  & 0.707                   & 0.743                  & 0.821                   \\
SFA \cite{fang2019selective}            & 100\%                 & 0.607                  & 0.700                   & 0.217                  & 0.297                   & 0.347                  & 0.469                   & 0.329                  & 0.467                   & 0.611                  & 0.723                   \\
PraNet \cite{pranet}                    & 100\%                 & 0.849                  & 0.899                   & 0.567                  & 0.628                   & 0.640                  & 0.709                   & 0.797                  & 0.871                   & 0.840                  & 0.898                   \\
MSNET \cite{zhao2021automatic}          & 100\%                 & \textbf{0.879}         & \textbf{0.921}          & 0.664                  & 0.719                   & 0.678                  & 0.755                   & 0.807                  & 0.869                   & 0.862                  & 0.907                   \\
Shallow Attention \cite{wei2021shallow} & 100\%                 & 0.859                  & 0.916                   & 0.654                  & 0.750                   & 0.670                  & 0.753                   & 0.815                  & 0.888                   & 0.847                  & 0.904                   \\ \hline
\multirow{3}{*}{Ours}                                    & 20\%                  & 0.816                  & 0.870                   & 0.702         & 0.777          & 0.669                  & 0.743                   & \textbf{0.836}                  & \textbf{0.904}          & 0.856                  & 0.912                   \\
                                                         & 40\%                  & 0.825                  & 0.883                   & \textbf{0.702}         & \textbf{0.777}          & 0.690                  & 0.757                   & 0.825                  & 0.895                   & 0.850                  & 0.909                   \\
                                                         & 60\%                  & 0.855                  & 0.902                   & 0.694                  & 0.772                   & \textbf{0.701}         & \textbf{0.767}          & 0.833                  & 0.899                   & \textbf{0.865}         & \textbf{0.916}          \\ \hline
\end{tabular}

\label{tab:compare_with_supervised}
\end{table*}

\section{Conclusion and Future Works}

This paper proposes a semi-supervised learning method that combines online pseudo label generation and momentum network. This method has been shown to generate pseudo labels without complicated settings. Our method achieves an average mDice of 85.33\% on a test set of 5 different data sets, CVC-300, CVC-ClinicDB, CVC-ColonDB, ETIS-LaribPolypDB, Kvasir-SEG. Especially, mDice on some datasets such as CVC-300, CVC-ColonDB, and ETIS-LaribPolypDB are outperformed with state-of-the-art supervised learning methods. This research is a promising direction in many practical applications because of the enormous amount of current and future unlabelled data, especially in support systems for diagnostics on medical images. We hope this research can promote the applications of semi-supervised learning in medical image segmentation problems in the future.

\section*{Acknowledgement}
This work is funded by Vingroup Innovation Foundation (VINIF) under project code VINIF.2020.DA17. This work is also partially supported by Sun-Asterisk Inc. We would like to thank our colleagues at Sun-Asterisk Inc for their advice and expertise. Without their support, this experiment would not have been accomplished.

\printbibliography
\end{document}